\def\year{2020}\relax
\documentclass[letterpaper]{article} 
\usepackage{aaai20}  
\usepackage{times}  
\usepackage{helvet} 
\usepackage{courier}  
\usepackage[hyphens]{url}  
\usepackage{graphicx} 
\usepackage{graphicx}  
\usepackage{multirow}
\usepackage{amsmath}
\usepackage{amsfonts}
\usepackage{subcaption}
\newcommand{\newcite}[1]{\citeauthor{#1}\shortcite{#1}}
\title{Graph Transformer for Graph-to-Sequence Learning\thanks{The work described in this paper is substantially supported by a grant from the Research Grant Council of the Hong Kong Special Administrative Region, China (Project Code: 14204418).}}
\author{Deng Cai \and Wai Lam \\
	The Chinese University of Hong Kong\\ 
	thisisjcykcd@gmail.com, wlam@se.cuhk.edu.hk 
}
\begin{document}
	\maketitle
	
	\begin{abstract}
		The dominant graph-to-sequence transduction models employ graph neural networks for graph representation learning, where the structural information is reflected by the receptive field of neurons. Unlike graph neural networks that restrict the information exchange between immediate neighborhood, we propose a new model, known as Graph Transformer, that uses explicit relation encoding and allows direct communication between two distant nodes. It provides a more efficient way for global graph structure modeling. Experiments on the applications of text generation from Abstract Meaning Representation (AMR) and syntax-based neural machine translation show the superiority of our proposed model. Specifically, our model achieves 27.4 BLEU on LDC2015E86 and 29.7 BLEU on LDC2017T10 for AMR-to-text generation, outperforming the state-of-the-art results by up to 2.2 points. On the syntax-based translation tasks, our model establishes new single-model state-of-the-art BLEU scores, 21.3 for English-to-German and 14.1 for English-to-Czech, improving over the existing best results, including ensembles, by over 1 BLEU.
	\end{abstract}
	\section{Introduction}
\begin{figure}[t]
		\centering
		\includegraphics[scale=0.28]{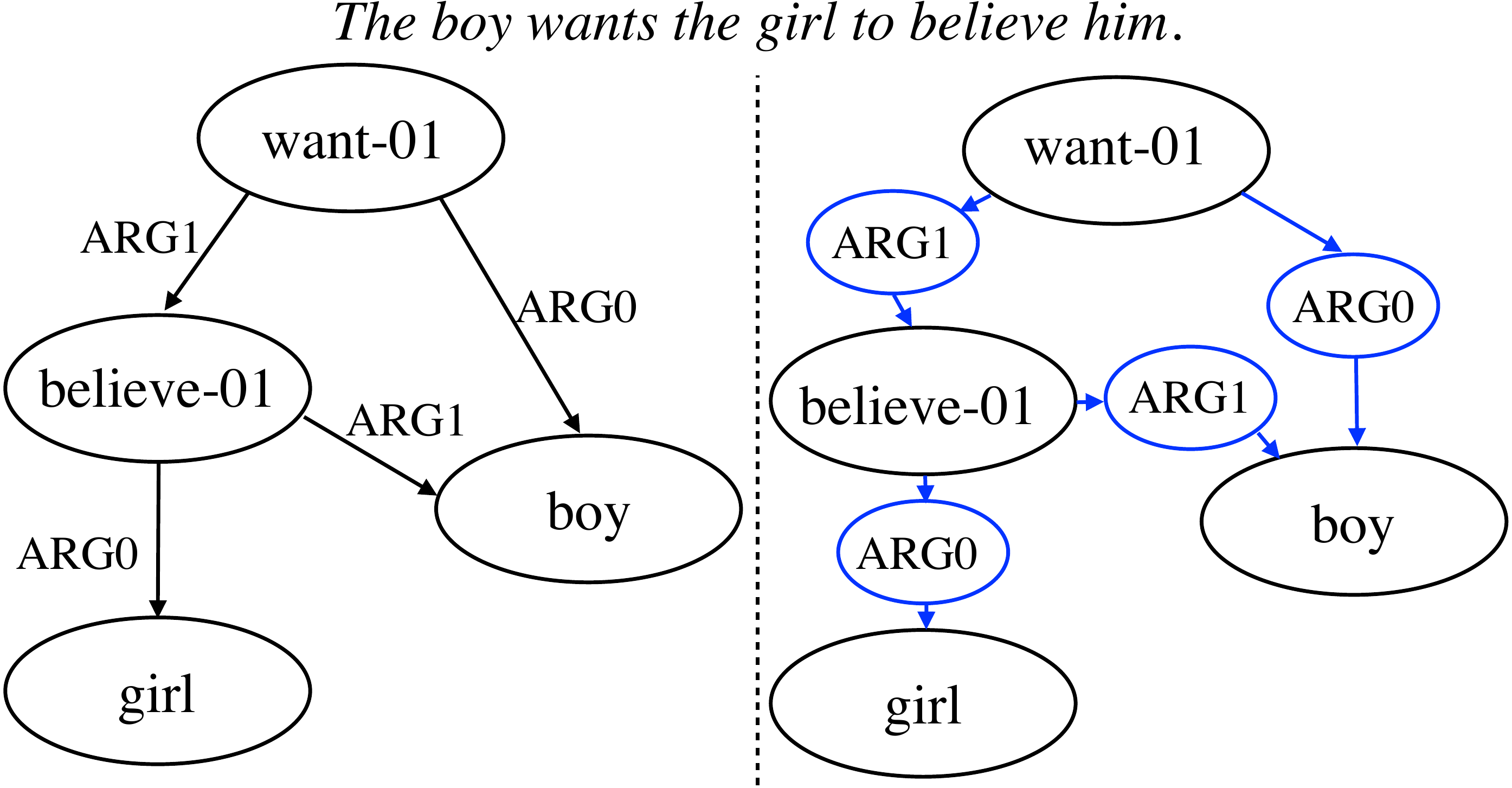}
		\caption{An AMR graph (left) for the reference sentence ``The boy wants the girl to believe him." and the corresponding Levi graph (right).}
		\label{example}
	\end{figure}

	Graphical structure plays an important role in natural language processing (NLP), they often serve as the central formalism for representing syntax, semantics, and knowledge. For example, most syntactic representations (e.g., dependency relation) are tree-based while most whole-sentence semantic representation frameworks (e.g., Abstract Meaning Representation (AMR) \cite{banarescu2013abstract}) encode sentence meaning as directed acyclic graphs. A range of NLP applications can be framed as the process of graph-to-sequence learning. For instance, text generation may involve realizing a semantic graph into a surface form \cite{liu-etal-2015-toward} and syntactic machine translation incorporates source-side syntax information for improving translation quality \cite{bastings-etal-2017-graph}. Fig. \ref{example} gives an example of AMR-to-text generation.
	
	While early work uses statistical methods or neural models after the linearization of graphs, graph neural networks (GNNs) have been firmly established as the state-of-the-art approaches for this task \cite{damonte-cohen-2019-structural,guo2019densely}. GNNs typically compute the representation of each node iteratively based on those of its adjacent nodes. This inherently local propagation nature precludes efficient global communication, which becomes critical at larger graph sizes, as the distance between two nodes exceeds the number of stacked layers. For instance, for two nodes staying $L$ hops away, at least $L$ layers will be needed in order to capture their dependencies. Furthermore, even if two distant nodes are reachable, the information may also be disrupted in the long journey \cite{xu2018graph2seq,guo2019densely}.
	
	To address the above problems, we propose a new model, known as Graph Transformer, which relies entirely on the multi-head attention mechanism \cite{vaswani2017attention} to draw global dependencies.\footnote{We note that the name \textit{Graph Transformer} was used in a recent work \cite{koncel-kedziorski-etal-2019-text}. However, it merely focuses on the relations between directly connected nodes as other graph neural networks.} Different to GNNs, the Graph Transformer allows direct modeling of dependencies between any two nodes without regard to their distance in the input graph. One undesirable consequence is that it essentially treats any graph as a fully connected graph, greatly diluting the explicit graph structure. To maintain a graph structure-aware view,  our proposed model introduces explicit relation encoding and incorporates it into the pairwise attention score computation as a dynamic parameter.
	
	Our treatment of explicit relation encoding also brings other side advantages compared to GNN-based methods. Previous state-of-the-art GNN-based methods use Levi graph transformation \cite{beck-etal-2018-graph,guo2019densely}, where two unlabeled edges are replacing one labeled edge that is present in the original graph. For example, in Fig. \ref{example}, the labeled edge  $\texttt{want-01}\stackrel{ARG1}{\longrightarrow}\texttt{believe-01}$ turns to be two unlabeled edges $\texttt{want-01}\stackrel{}{\longrightarrow}\texttt{ARG1}$ and $\texttt{ARG1}\stackrel{}{\longrightarrow}\texttt{believe-01}$. Since edge labels are represented as nodes, they end up sharing the same semantic space, which is not ideal as nodes and edges are typically different elements. In addition, the Levi graph transformation at least doubles the number of representation vectors. which will introduce more complexity for the decoder-side attention mechanism \cite{bahdanau2015neural} and copy mechanism \cite{gu-etal-2016-incorporating,see-etal-2017-get}. Through explicit and separate relation encoding, our proposed Graph Transformer inherently avoids these problems.
	
	Experiments show that our model is able to achieve better performance for graph-to-sequence learning tasks for natural language processing. For the AMR-to-text generation task, our model surpasses the current state-of-the-art neural methods trained on LDC2015E86 and LDC2017T10 by 1.6 and 2.2 BLEU points, respectively. For the syntax-based neural machine translation task, our model is also consistently better than others, even including ensemble systems, showing the effectiveness of the model on a large training set. In addition, we give an in-depth study of the source of improvement gain and the internal workings of the proposed model.
	\section{Related Work}
	Early research efforts for graph-to-sequence learning use specialized grammar-based methods. \newcite{flanigan-etal-2016-generation} split input graphs to trees and uses a tree-to-string transducer. \newcite{song-etal-2016-amr} recast generation as a traveling salesman problem. \newcite{jones-etal-2012-semantics} leverage hyperedge replacement grammar and \newcite{song-etal-2017-amr} use a synchronous node replacement grammar. More recent work employs more general approaches, such as phrase-based machine translation model \cite{pourdamghani2016generating} and neural sequence-to-sequence methods \cite{konstas-etal-2017-neural} after linearizing input graphs. Regarding AMR-to-text generation, \newcite{cao-clark-2019-factorising} propose an interesting idea that factorizes text generation through syntax. One limitation of sequence-to-sequence models, however, is that they require serialization of input graphs, which inevitably incurs the obstacle of capturing graph structure information.
	
	An emerging trend has been directly encoding the graph with different variants of graph neural networks, which in common stack multiple layers that restrict the update of node representation based on a first-order neighborhood but use different information passing schemes. Some borrow the ideas from recurrent neural networks (RNNs), e.g, \newcite{beck-etal-2018-graph} use gated graph neural network \cite{li2016gated} while \newcite{song-etal-2018-graph} introduce LSTM-style information aggregation. Others apply convolutional neural networks (CNNs), e.g., \newcite{bastings-etal-2017-graph};\newcite{damonte-cohen-2019-structural};\newcite{guo2019densely} utilize graph convolutional neural networks \cite{kipf2017semi}. \newcite{koncel-kedziorski-etal-2019-text} update vertex information by attention over adjacent neighbors. Furthermore, \newcite{guo2019densely} allow the information exchange across different levels of layers. \newcite{damonte-cohen-2019-structural} systematically compare different encoders and show the advantages of graph encoder over tree and sequential ones. The contrast between our model and theirs is reminiscent of the contrast between the self-attention network (SAN) and CNN/RNN.
	
	For sequence-to-sequence learning, the SAN-based Transformer model \cite{vaswani2017attention} has been the \textit{de facto} approach for its empirical successes. However, it is unclear on the adaptation to graphical data and its performance. Our work is partially inspired by the introduction of relative position embedding \cite{shaw-etal-2018-self,dai-etal-2019-transformer} in sequential data. However, the extension to graph is nontrivial since we need to model much more complicated relation instead of mere visual distance. To the best of our knowledge, the Graph Transformer is the first graph-to-sequence transduction model relying entirely on self-attention to compute representations.
	\begin{figure*}[t]
		\centering
		\includegraphics[scale=0.44]{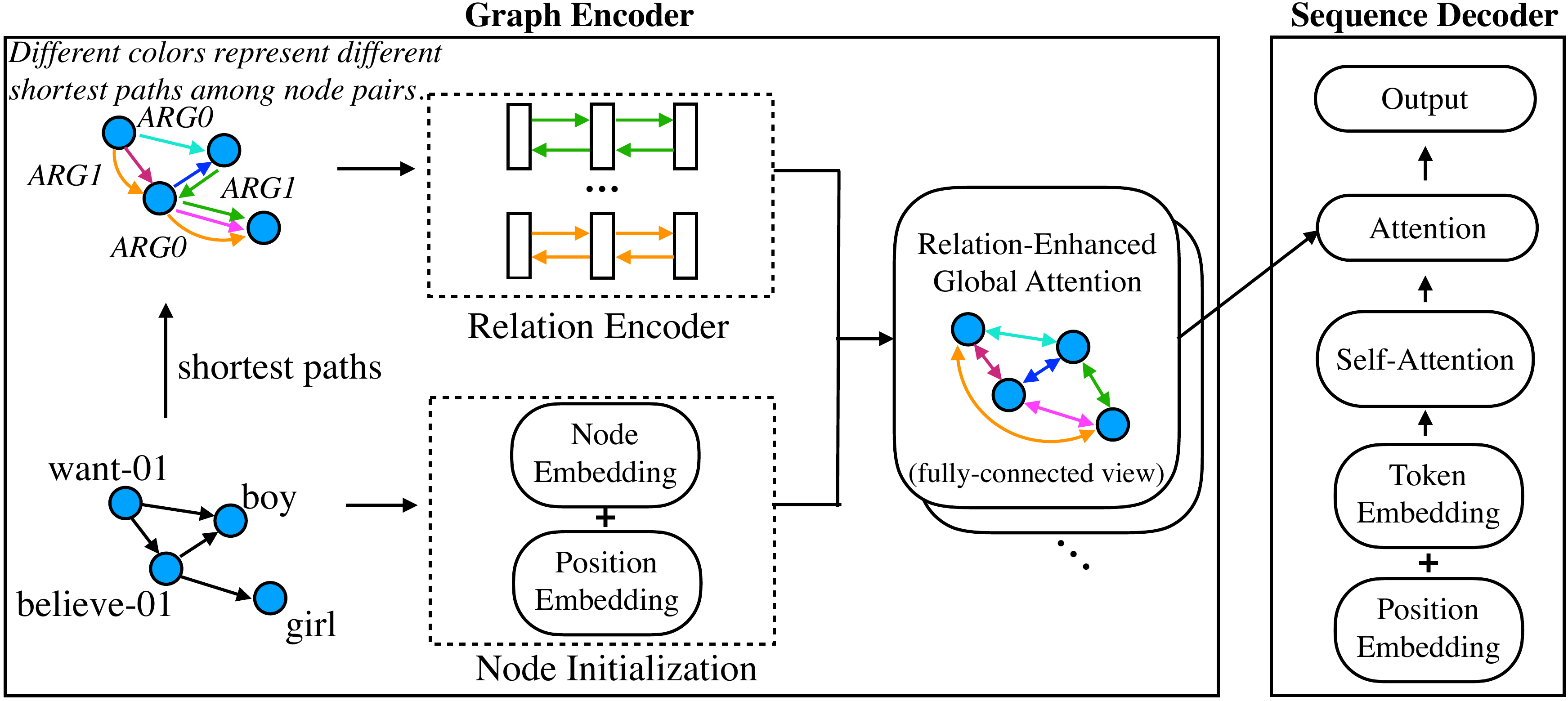}
		\caption{An overview of our proposed model.}
		\label{arch}
	\end{figure*}
	\section{Background of Self-Attention Network}
	The Transformer introduced by \newcite{vaswani2017attention} is a sequence-to-sequence neural architecture  originally used for neural machine translation. It employs self-attention network (SAN) for implementing both the encoder and the decoder. The encoder consists of multiple identical blocks, of which the core is multi-head attention. The multi-head attention consists of $H$ attention heads, and each of them learns a distinct attention function. Given a source vector $x\in\mathbb{R}^{d_x}$ and a set of context vectors $\{y_1, y_2, \ldots, y_m\}$ with the same dimension $d_x$ or in short $y_{1:m}$, for each attention head, $x$ and $y_{1:m}$ are transformed into distinct query and value representations. The attention score is computed as the dot-product between them.
	\begin{align*}
		f(x, y_i) =(W_qx)^TW_k y_i
	\end{align*}
	where $W_q, W_k \in \mathbb{R}^{d_z\times d_x}$ are trainable projection matrices. The attention scores are scaled and normalized by a softmax function to compute the final attention output $attn$.
	\begin{align*}
		&a_i = \frac{\exp(f(x, y_i)/ \sqrt{d_z})}{\sum_{j=1}^m \exp(f(x, y_i))/ \sqrt{d_z})}\\
		&attn = \sum_{i=1}^m a_i W_v{y_i}
	\end{align*}
	where $a \in \mathbb{R}^m$ is the attention vector (a distribution over all input $y_{1:m}$), $W_v\in\mathbb{R}^{d_z \times d_x}$ is a trainable projection matrix. Finally, the outputs of all attention heads are concatenated and projected to the original dimension of $x$, followed by feed-forward layers, residual connection, and layer normalization.\footnote{We refer interesting readers to \newcite{vaswani2017attention} for more details.} For brevity, we will denote the whole procedure described above as a single function $\textsc{ATT}(x,y_{1:m})$.
	
	For an input sequence $x_{1:n}$, the SAN-based encoder computes the vector representations iteratively by  $x^L_i  = \textsc{ATT}(x^L_i,x^{L-1}_{1:n})$, where $L$ is the total number of blocks and $x^0_{1:n}$ are word embeddings. In this way, a representation is allowed to build a direct relationship with another long-distance representation. To feed the sequential order information, the deterministic or learned position embedding \cite{vaswani2017attention} is introduced to expose the position information to the model, i.e., $x^0_{i}$ becomes the sum of the corresponding word embedding and the position embedding for $i$.
	
	The aforementioned treatment of SAN on sequential data can be drawn a close resemblance to graph neural networks by regarding the token sequence as an unlabeled fully-connected graph (each token as a node) and taking the multi-head attention mechanism as a specific message-passing scheme. Such view on the relationship between SAN and graph neural networks inspires our work. 
	\section{Graph Transformer}
	\subsection{Overview}
	For a graph with $n$ nodes, previous graph neural networks compute the node representation $v_i$ as a function of the input node $i$ and all its first-order neighborhoods $N(i)$. The graph structure is implicitly reflected by the receptive field of each node representation. This local communication design, however, could be inefficient for long-distance information exchange. We introduce a new model, known as Graph Transformer, which provides an aggressively different paradigm that enables relation-aware global communication.
	
	The overall framework is shown in Fig. \ref{arch}. The most important characteristic of the Graph Transformer is that it has a fully-connected view on arbitrary input graphs. A node is able to directly receive and send information to another node no matter whether they are directly connected or not. These operations are achieved by our proposed extension to the original multi-head attention mechanism, the relation-enhanced global attention mechanism described below. In a nutshell, the relationship between any node pair is depicted as the shortest relation path between them. These pairwise relation paths are fed into a relation encoder for distributed relation encoding. The node vectors are initialized as the sum of the node embedding and absolute position embeddings. Multiple blocks of global attention network are then stacked to compute the final node representations. At each block, a node vector is updated based on all other node vectors and the corresponding relation encodings. The resulted node vectors at the last block are fed to the sequence decoder for sequence generation.
	\subsection{Graph Encoder}
	Our graph encoder is responsible for transforming an input graph into a set of corresponding node embeddings. To apply global attention on a graph, the central problem is how to maintain the topological structure of the graph while allowing fully-connected communication. To this end, we propose relation-enhanced global attention mechanism, which is an extension of the vanilla multi-head attention. Our idea is to incorporate explicit relation representation between two nodes into their representation learning. Recall that, in the standard multi-head attention, the attention score between the element $x_i$ and the element $x_j$ is simply the dot-product of their query vector and key vector respectively:
	\begin{align}
		\begin{split}
			s_{ij} & = f(x_i, x_j) \\
			& =x_i W_q^TW_k x_j
		\end{split}
		\label{datt}
	\end{align}
	Suppose we have learned a vector representation for the relationship $r_{ij}$, which we will refer as relation encoding, between the node $i$ and the node $j$. Following the idea of relative position embedding \cite{shaw-etal-2018-self,dai-etal-2019-transformer}, we propose to compute the attention score as follows:
	\begin{align}
		& [r_{i\to j}; r_{j\to i}] = W_r r_{ij}
		\label{r_split}
	\end{align}
	where we first split the relation encoding $r_{ij}$ into the forward relation encoding $r_{i\to j}$ and the backward relation encoding $r_{j\to i}$. Then we compute the attention score based on both the node representations and their relation representation: 
	\begin{align}
		\begin{split}
			s_{ij} &= g(x_i, x_j, r_{ij})\\
			& = (x_i + r_{i\to j})W_q^TW_k(x_j + r_{j\to i})\\
			&= \underbrace{x_iW_q^TW_kx_j}_{(a)} +  \underbrace{x_iW_q^TW_kr_{j\to i}}_{(b)} \\
			&+ \underbrace{r_{i\to j}W_q^TW_kx_j}_{(c)} + \underbrace{r_{i\to j}W_q^TW_kr_{j\to i}}_{(d)}
		\end{split}
		\label{ratt}
	\end{align}
	Each term in Eq (\ref{ratt}) corresponds to some intuitive meaning according to their formalization. The term (a) captures purely content-based addressing, which is the original term in vanilla attention mechanism. The term (b) represents a source-dependent relation bias. The term (c) governs a target-dependent relation bias. The term (d) encodes the universal relation bias. Our formalization provides a principled way to model the element-relation interactions. In comparison, it has broader coverage than \newcite{shaw-etal-2018-self} in terms of additional terms (c) and (d), and than \newcite{dai-etal-2019-transformer} in terms of the extra term (c) respectively. More importantly, previous methods only model the relative position in the context of sequential data, which merely adopts the immediate embeddings of the relative positions (e.g, $-1, +1$). To depict the relation between two nodes in a graph, we utilize a shortest-path based approach as described below.
	\subsubsection{Relation Encoder}
	Conceptually, the relation encoding gives the model a global guidance about how information should be gathered and distributed, i.e., where to attend. For most graphical structures in NLP, the edge label conveys direct relationship between adjacent nodes (e.g., the semantic role played by concept-to-concept, and the dependency relation between two words). We extend this one-hop relation definition into multi-hop relation reasoning for characterizing the relationship between two arbitrary nodes. For example, in Fig \ref{example}, the shortest path from the concept \texttt{want-01} to \texttt{girl} is `` $\texttt{want-01}\stackrel{ARG1}{\longrightarrow}\texttt{believe-01}\stackrel{ARG0}{\longrightarrow}\texttt{girl}$", which conveys that \texttt{girl} is the object of the \textit{wanted} action. Intuitively, the shortest path between two nodes gives the closest and arguably the most important relationship between them. Therefore, we propose to use the shortest paths (relation sequence) between two nodes to characterize their relationship.\footnote{For the case that there are multiple shortest paths, we randomly sample one during training and take the averaged representation during testing.} Following the sequential nature of the relation sequence, we employs recurrent neural networks with Gated Recurrent Unit (GRU) \cite{cho2014learning} for transforming relation sequence into a distributed representation. Formally,  we represent the shortest relation path $sp_{i \to j} = [e(i, k_1), e(k_1, k_2), \ldots, e(k_n, j)]$ between the node $i$ and the node $j$, where $e(\cdot, \cdot)$ indicates the edge label and $k_{1:n}$ are the relay nodes. We employ bi-directional GRUs for sequence encoding:
	\begin{align*}
		\overrightarrow{s_t} &= \text{GRU}_f( \overrightarrow{s_{t-1}}, sp_t) \\
		\overleftarrow{s_t} &= \text{GRU}_b(\overleftarrow{s_{t+1}}, sp_t)
	\end{align*}
	The last hidden states of the forward GRU network and the backward GRU networks are concatenated to form the final relation encoding $r_{ij} = [ \overrightarrow{s_n}; \overleftarrow{s_0}]$.
	\subsubsection{Bidirectionality} Though in theory, our architecture can deal with arbitrary input graphs, the most widely adopted graphs in the real problems are directed acyclic graphs (DAGs). This implies that the node embedding information will be propagated in one pre-specified direction. However, the reverse direction informs the equivalent information flow as well. To facilitate communication in both directions, we add reverse edges to the graph. The reverse edge connects the same two nodes as the original edge but in a different direction and with a reversed label. For example, we will draw a virtual edge $\texttt{believe-01}\stackrel{RARG1}{\longrightarrow}\texttt{want-01}$ according to the original edge $\texttt{want-01}\stackrel{ARG1}{\longrightarrow}\texttt{believe-01}$. For convenience, we also introduce self-loop edges for each node. These extra edges have specific labels, hence their own parameters in the network. We also introduce an extra global node into every graph, who has a direct edge to all other nodes with the special label $global$. The final representation $x_{global}$ of the global node serves as a whole graph representation.
	\subsubsection{Absolute Position} Besides pairwise relationship, some absolute positional information can also be beneficial. For example, the root of an AMR graph serves as a rudimentary representation of the overall focus, making the minimum distance from the root node partially reflect the importance of the corresponding concept in the whole-sentence semantics. The sequence order of tokens in a dependency tree also provides complementary information to dependency relations. In order for the model to make use of the absolute positions of nodes, we add the positional embeddings to the input embeddings at the bottom of the encoder stacks. For example, \texttt{want-01} in Fig \ref{example} is the root node of the AMR graph, so its index should be 0. Notice we denote the index of the global node as $0$ as well.
	\begin{table*}[t]
		\small
		\centering
		\begin{tabular}{c|c|c|c|c|c|c|c|c}
			Dataset & \#train & \#dev & \#test & \#edge types & \#node types & avg \#nodes& avg \#edges & avg diameter\\
			\hline
			LDC2015E86 & 16,833 & 1,368 & 1,371 & 113& 18735 &17.34&17.53&6.98\\
			LDC2017T10 & 36,521 & 1,368 & 1,371  & 116& 24693 &14.51&14.62&6.15\\
			\hline
			English-Czech & 181,112 & 2,656 & 2,999 &46&78017&23.18&22.18&8.36\\
			English-German & 226,822 & 2,169 & 2,999 &46&87219&23.29&22.29&8.42
		\end{tabular}
		\caption{Data statistics of all four datasets. \#train/dev/test indicates the number of instances in each set, avg \#nodes/edges/diameter represents the averaged value of nodes/edge/diameter size of a graph.}
		\label{data}
	\end{table*}
	\subsection{Sequence Decoder}
	Our sequence decoder basically follows the same spirit of the sequential Transformer decoder. The decoder yields the natural language sequence by calculating a sequence of hidden states sequentially. One distinct characteristic is that we use the global graph representation $x_{global}$ for initializing the hidden states at each time step. The hidden state $h_t$ at each time step $t$ is then updated by interleaving multiple rounds of attention over the output of the encoder (node embeddings) and attention over previously-generated tokens (token embeddings). Both are implemented by the multi-head attention mechanism. $x_{global}$ is removed when performing the sequence-to-graph attention.
	\subsubsection{Copy mechanism}
	To address the data sparsity issue in token prediction, we include a copy mechanism \cite{gu-etal-2016-incorporating} in similar spirit to most recent works. Concretely, a single-head attention is computed based on the decoder state $h_t$ and the node representation $x_{1:n}$, where $a_t^i$ denotes the attention weight of the node $v_i$ in the current time step $t$. Our model can either directly copy the type name of a node (node label) or generate from a pre-defined vocabulary $V$. Formally, the prediction probability of a token $y$ is given by:
	\begin{align*}
		P(y|h_t) = P(gen|h_t) gen(y|h_t) +P(copy|h_t) \sum_{i \in S(y)} a_t^i
	\end{align*}
	where $S(y)$ is the set of nodes that have the same surface form as $y$. $P(gen|h_t)$ and $P(copy|h_t)$ are computed by a single layer neural network with softmax activation, and $gen(y|h_t) = \exp({w_y}^T h_t)/ \sum_{y'\in V} \exp({w_y'}^T h_t)$, where $w_y$ (for $y \in V$) denotes the model parameters. The copy mechanism facilitates the generation of dates, numbers, and named entities in both AMR-to-text generation and machine translation tasks in experiments.
	\begin{table}[t]
		\centering
		\small
		\begin{tabular}{c|c|c}
			model component&  hyper-parameter& value\\
			\hline
			\multirow{4}{*}{char-level CNN} & number of filters & 256 \\
			& width of filters & 3 \\
			& char embedding size & 32 \\
			& final hidden size & 128 \\
			\hline
			\multirow{2}{*}{Embeddings} & node embedding size & 300 \\
			&edge embedding size& 200\\
			&token embedding size&300\\
			\hline
			\multirow{3}{*}{Multi-head attention} & number of heads & 8 \\
			& hidden state size & 512 \\
			& feed-forward hidden size  & 1024
		\end{tabular}
		\caption{Hyper-parameters settings.}
		\label{hyper}
	\end{table}
	\begin{table*}[t]
		\small
		\centering
		\begin{tabular}{c|c|c|c|c|c|c}
			\multirow{2}{*}{Model}& \multicolumn{3}{c|}{LDC2015E86} & \multicolumn{3}{c}{LDC2017T10}\\
			\cline{2-7}
			& \textsc{BLEU} &  \textsc{chrF++} &  \textsc{Meteor}& \textsc{BLEU} &  \textsc{chrF++} &  \textsc{Meteor}\\
			\hline
			\newcite{song-etal-2016-amr}$\dag$  & 22.4 &-&-&-&-&-\\
			\newcite{flanigan-etal-2016-generation}$\dag$   &23.0&-&-&-&-&-\\
			\newcite{pourdamghani2016generating}$\dag$ & 26.9 &-&-&-&-&-\\
			\newcite{song-etal-2017-amr}$\dag$ &25.6&-&-&-&-&-\\
			\hline
			\newcite{konstas-etal-2017-neural}   &22.0&-& -& -&-&-\\
			\newcite{cao-clark-2019-factorising}$\ddag$ &23.5 &-&-&26.8&-&-\\
			\hline
			\newcite{song-etal-2018-graph} & 23.3& -&-&24.9&-&-\\
			\newcite{beck-etal-2018-graph}& -& -&-&23.3&50.4&\\
			\newcite{damonte-cohen-2019-structural} &24.4&-&23.6&24.5&-&24.1\\
			\newcite{guo2019densely} &25.7&54.5$^*$&31.5$^*$&27.6&57.3&34.0$^*$ \\
			\hline
			Ours &\textbf{27.4}&\textbf{56.4}&\textbf{32.9}&\textbf{29.8}&\textbf{59.4}&\textbf{35.1}
		\end{tabular}
		\caption{Main results on AMR-to-text generation. Numbers with $^*$ are from the contact from the authors. - denotes that the result is unknown because it is not provided in the corresponding paper.}
		\label{main-amr}
	\end{table*}
	\begin{table*}[t]
		\centering
		\small
		\begin{tabular}{c|c|c|c|c|c}
			\multirow{2}{*}{Model}& \multirow{2}{*}{Type} &\multicolumn{2}{c|}{English-German} & \multicolumn{2}{c}{English-Czech}\\
			\cline{3-6}
			&&\textsc{BLEU} & \textsc{chrF++} & \textsc{BLEU} &  \textsc{chrF++} \\
			\hline
			\newcite{bastings-etal-2017-graph} &Single &16.1&-&9.6&-\\
			\newcite{beck-etal-2018-graph} &Single & 16.7& 42.4&9.8&33.3\\
			\newcite{guo2019densely} &Single &19.0&44.1&12.1&37.1 \\
			\hline
			\newcite{beck-etal-2018-graph} &Ensemble & 19.6& 45.1&11.7&35.9\\
			\newcite{guo2019densely} &Ensemble &20.5&45.8&13.1& 37.8\\
			\hline
			Ours &Single&\textbf{21.3}&\textbf{47.9}&\textbf{14.1}&\textbf{41.1}
		\end{tabular}
		\caption{Main results on syntax-based machine translation.}
		\label{main-nmt}
	\end{table*}
	\begin{figure*}[t]
		\centering
		\begin{subfigure}{0.32\textwidth}
		\includegraphics[width=0.99\linewidth]{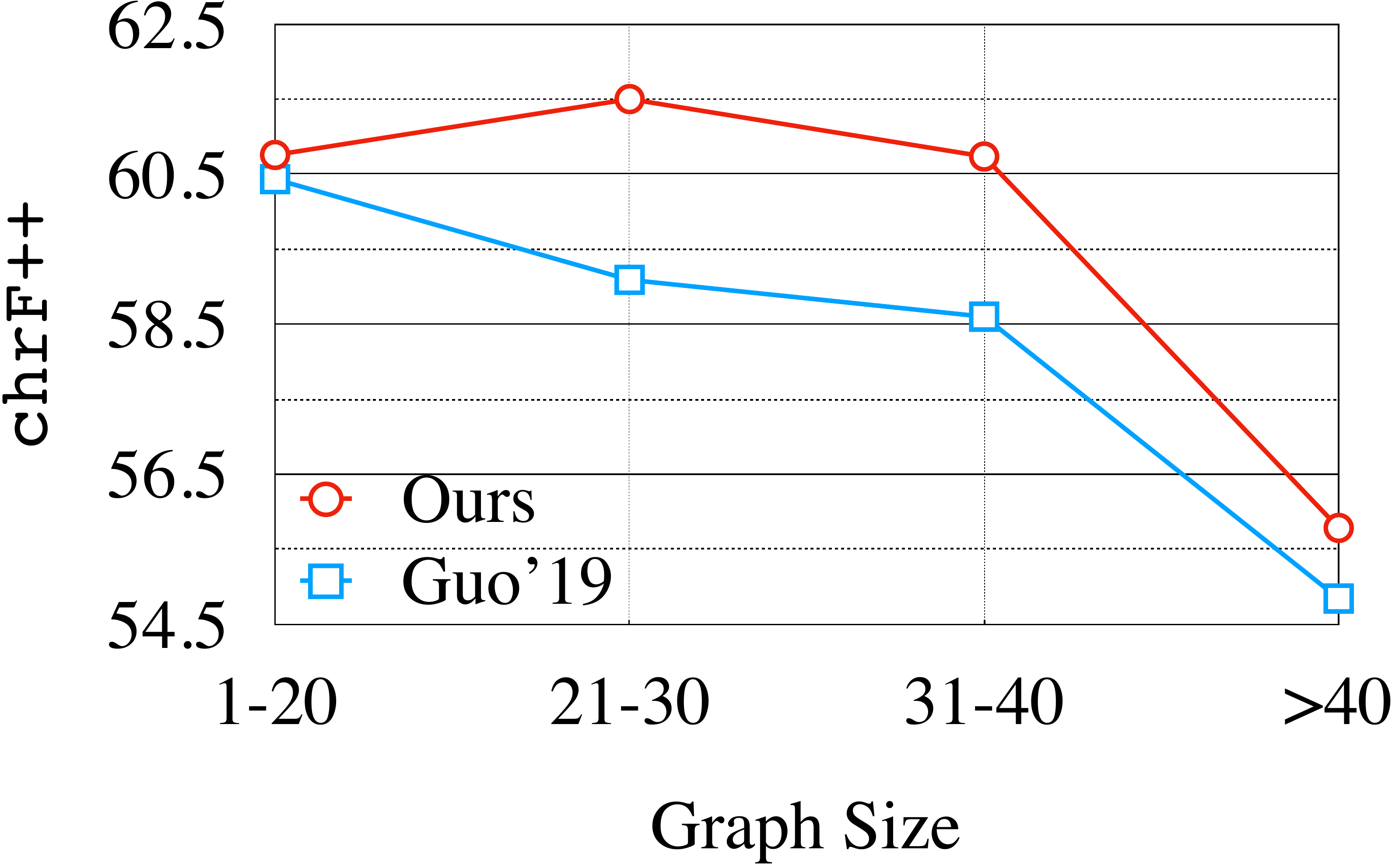}
				\caption{}
			\label{graph_size}
		\end{subfigure}
			\begin{subfigure}{0.32\textwidth}
		\includegraphics[width=0.99\linewidth]{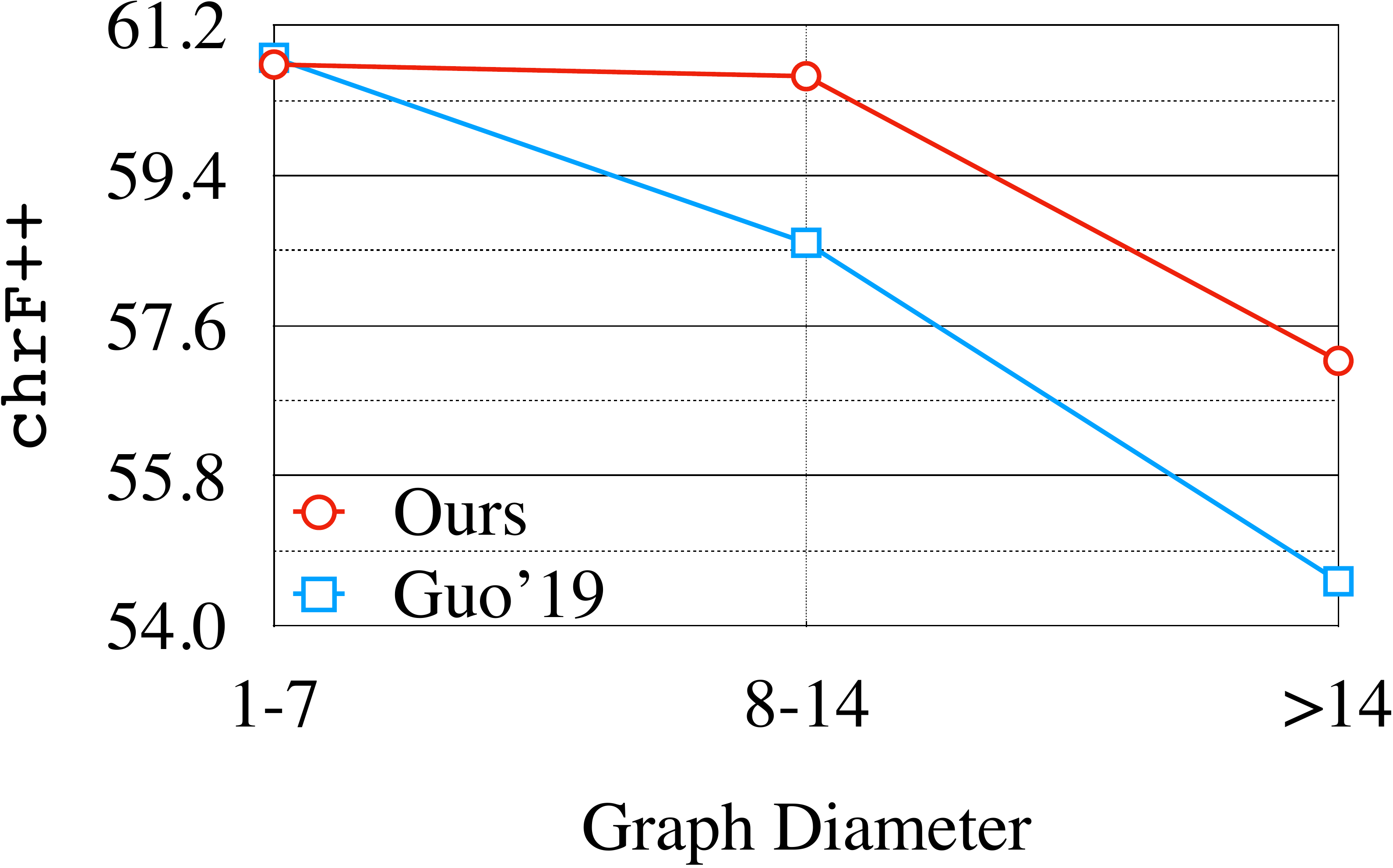}
			\caption{}
		\label{graph_diameter}
	\end{subfigure}
		\begin{subfigure}{0.32\textwidth}
	\includegraphics[width=0.99\linewidth]{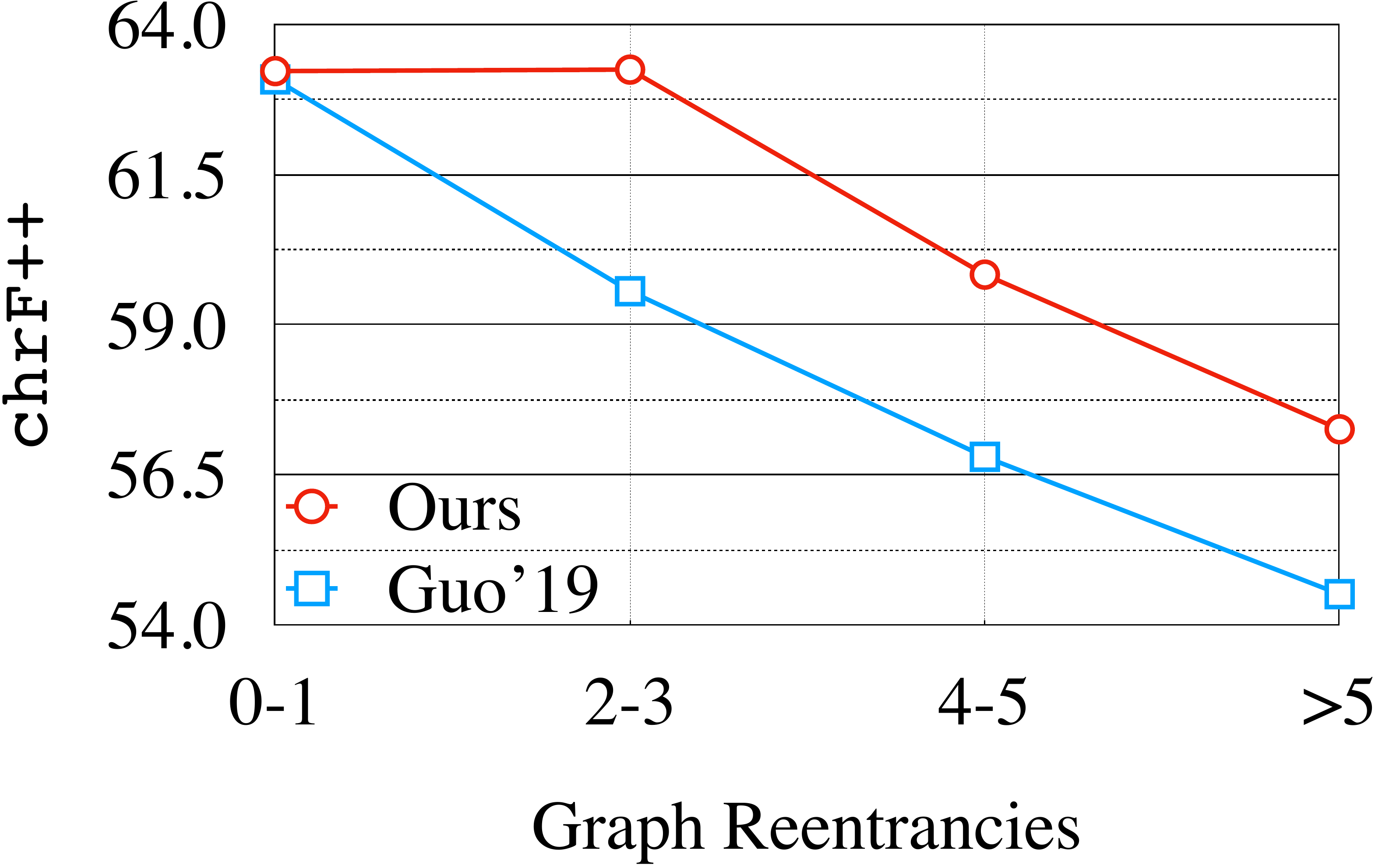}
			\caption{}
			\label{graph_re}
\end{subfigure}
		\caption{\textsc{chrF++} scores with respect to (a) the graph size, (b) the graph diameter, and (c) the the number of reentrancies.}
	\end{figure*} 
	\section{Experiments}
	\label{sec-exp}
	We assess the effectiveness of our models on two typical graph-to-sequence learning tasks, namely AMR-to-text generation and syntax-based machine translation (MT). Following previous work, the results are mainly evaluated by \textsc{BLEU} \cite{papineni2002bleu} and \textsc{chrF++} \cite{popovic2017chrf++}. Specifically, we use case-insensitive scores for AMR and case-sensitive BLEU scores for MT.
	\subsection{AMR-to-text Generation}
	Our first application is language generation from AMR, a semantic formalism that represents sentences as rooted DAGs \cite{banarescu2013abstract}. For this AMR-to-text generation task, we use two benchmarks, namely the LDC2015E86 dataset and the LDC2017T10 dataset. The first block of Table \ref{data} shows the statistics of the two datasets. Similar to \newcite{konstas-etal-2017-neural}, we apply entity simplification and anonymization in the preprocessing steps and restore them in the postprocessing steps.
	
	The graph encoder uses randomly initialized node embeddings as well as the output from a learnable CNN with character embeddings as input. The sequence decoder uses randomly initialized token embeddings and another char-level CNN. Model hyperparameters are chosen by a small set of experiments on the development set of LDC2017T10. The detailed settings are listed in Table \ref{hyper}. During testing, we use a beam size of $8$ for generating graphs. To mitigate overfitting, we also apply dropout \cite{srivastava2014dropout} with the drop rate of $0.2$ between different layers. We use a special UNK token to replace the input node tag with a rate of $0.33$. Parameter optimization is performed with the Adam optimizer \cite{kingma2014adam} with $ \beta_1= 0.9$ and $beta_2 = 0.999$. The same learning rate schedule of \newcite{vaswani2017attention} is adopted in our experiments.\footnote{Code available at \url{https://github.com/jcyk/gtos}.}  For computation efficiency, we gather all distinct shortest paths in a training/testing batch, and encode them into vector representations by the recurrent relation encoding procedure as described above.\footnote{This strategy reduces the number of related sequences to encode from $O(mn^2)$ to a stable number when a large batch size $m$ is used.} 
	
	We run comparisons on systems without ensembling nor additional silver data. Specifically, the comparison methods can be grouped into three categories: (1) feature-based statistical methods \cite{song-etal-2016-amr,pourdamghani2016generating,song-etal-2017-amr,flanigan-etal-2016-generation}; (2) sequence-to-sequence neural models \cite{konstas-etal-2017-neural,cao-clark-2019-factorising}, which use linearized graphs as inputs; (3) recent works using different variants of graph neural networks for encoding graph structures directly \cite{song-etal-2018-graph,beck-etal-2018-graph,damonte-cohen-2019-structural,guo2019densely}. The results are shown in Table \ref{main-amr}. For both datasets, our approach substantially outperforms all previous methods. On the LDC2015E86 dataset, our method achieves a BLEU score of 27.4, outperforming previous best-performing neural model \cite{guo2019densely} by a large margin of 2.6 BLEU points. Also, our model becomes the first neural model that surpasses the strong non-neural baseline established by \newcite{pourdamghani2016generating}. It is worth noting that those traditional methods marked with $\dag$ train their language models on the external Gigaword corpus, thus they possess an additional advantage of extra data. On the LDC2017T10 dataset, our model establishes a new record BLEU score of 29.8, improving over the state-of-the-art sequence-to-sequence model \cite{cao-clark-2019-factorising} by 3 points and the state-of-the-art GNN-based model \cite{guo2019densely} by 2.2 points. The results are even more remarkable since the model of \newcite{cao-clark-2019-factorising} (marked with $\ddag$) uses constituency syntax from an external parser. Similar phenomena can be found on the additional metrics of \textsc{chrF++} and \textsc{Meteor} \cite{denkowski:lavie:meteor-wmt:2014}. Those results suggest that current graph neural networks cannot make full use of the AMR graph structure, and our Graph Transformer provides a promising alternative.
	\subsection{Syntax-based Machine Translation}
	Our second evaluation is syntax-based machine translation, where the input is a source language dependency syntax tree and the output is a plain target language string. We employ the same data and settings from \newcite{bastings-etal-2017-graph}. Both the English-German and the English-Czech datasets from the WMT16 translation task.\footnote{\url{http://www.statmt.org/wmt16/translation-task.html.}} The English sentences are parsed after tokenization to generate the dependency trees on the source side using SyntaxNet \cite{alberti2017syntaxnet}.\footnote{\url{https://github.com/tensorflow/models/ tree/master/syntaxnet}} On the Czech and German sides, texts are tokenized using the Moses tokenizer.\footnote{\url{https://github.com/moses-smt/mosesdecoder.}} Byte-pair encodings \cite{sennrich-etal-2016-neural} with 8,000 merge operations are used to obtain subwords. The second block of Table \ref{data} shows the statistics for both datasets. For model configuration, we just re-use the settings obtained in our AMR-to-text experiments.
	
	Table \ref{main-nmt} presents the results with comparison to existing methods. On the English-to-German translation task, our model achieves a BLEU score of 41.0, outperforming all of the previously published single models by a large margin of 2.3 BLEU score. On the  English-to-Czech translation task, our model also outperforms the best previously reported single models by an impressive margin of 2 BLEU points. In fact, our single model already outperforms previous state-of-the-art models that use ensembling. The advantages of our method are also verified by the metric \textsc{chrF++}.
	
	An important point about these experiments is that we did not tune the architecture: we simply employed the same model in all experiments, only adjusting the batch size for different dataset size. We speculate that even better results would be obtained by tuning the architecture to individual tasks. Nevertheless, we still obtained improved performance over previous works, underlining the generality of our model.
	\section{More Analysis}
	The overall scores show a great advantage of the Graph Transformer over existing methods, including the state-of-the-art GNN-based models. However, they do not shed light into how this is achieved. In order to further reveal the source of performance gain, we perform a series of analyses based on different characteristics of graphs. For those analyses, we use sentence-level \textsc{chrF++} scores, and take the macro average of them when needed. All experiments are conducted with the test set of LDC2017T10.
	\begin{figure}[t]
		\centering
		\includegraphics[scale=0.41]{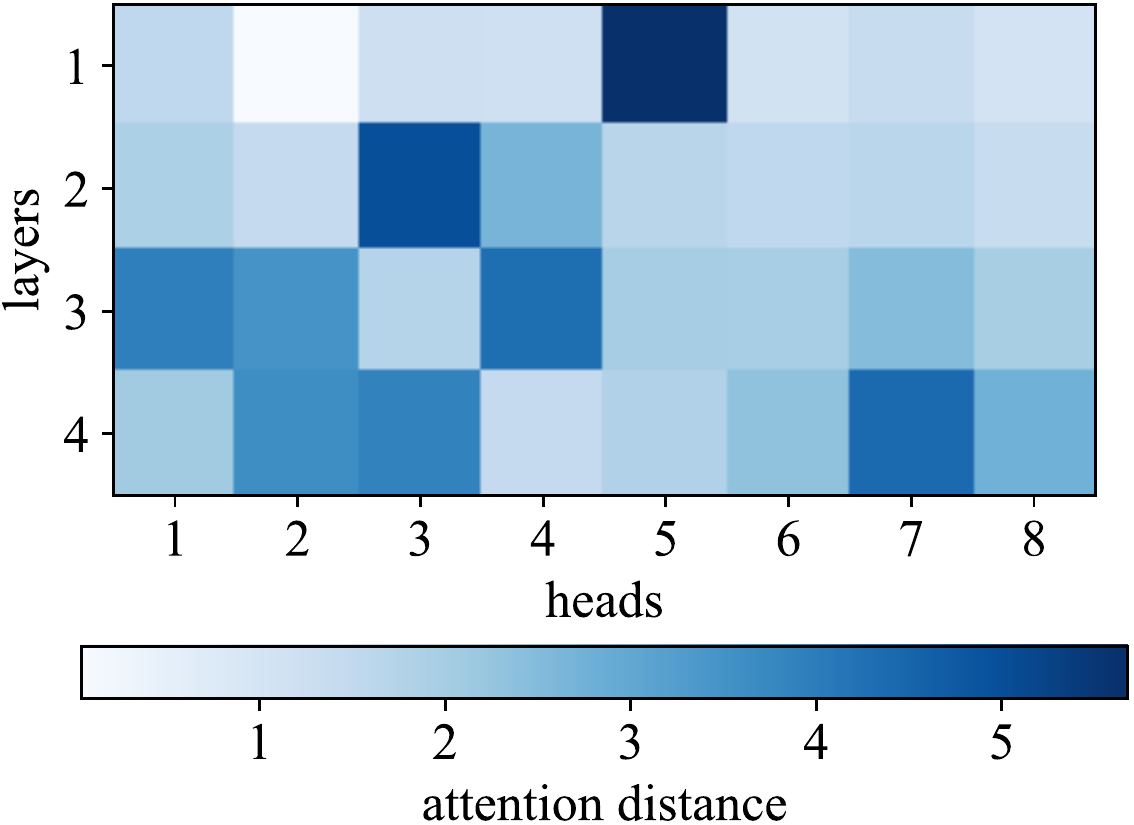}
		\caption{The average distance for maximum attention for each head.}
		\label{head}
	\end{figure} 
	\subsubsection{Graph Size} To assess the model's performance for different sizes of graphs, we group graphs into four classes and show the curves of \textsc{chrF++} scores in Figure \ref{graph_size}. The results are presented with the contrast with the state-of-the-art GNN-based model of \newcite{guo2019densely}, denoted as Guo'19. As seen, the performance of both models decreases as the graph size increases. It is expected since a larger graph often contains more complex structure and the interactions between graph elements are more difficult to capture. The gap between ours and Guo'19 becomes larger for relatively larger graphs while for small graphs, both models give similar performance. This result demonstrates that our model has better ability for dealing with complicated graphs. As for extremely large graphs, the performance of both models have a clear drop, yet ours is still slightly better. 
	\subsubsection{Graph Diameter} We then study the impact of graph diameter.\footnote{The diameter of a graph is defined as the length of the longest shortest path between two nodes.} Graphs with large diameters have interactions between two nodes that appear distant from each other. We conjecture that it will cause severe difficulties for GNN-based models because they solely rely on local communication. Figure \ref{graph_diameter} confirms our hypothesis, as the curve of the GNN-based model shows a clear slope. In contrast, our model has more stable performance, and the gap between the two curves also illustrates the superiority of our model on featuring long-distance dependencies.
	\subsubsection{Number of Reentrancies} We also study the ability for handling the reentrancies, where the same node has multiple parent nodes (or the same concept participates in multiple relations for AMR). The recent work \cite{damonte-cohen-2019-structural} has identified reentrancies as one of the most difficult aspects of AMR structure. We bin the number of reentrancies occurred in a graph into four classes and plot Fig. \ref{graph_re}. It can be observed that the gap between the GNN-based model and the Graph transformer becomes noticeably wide when there are more than one reentrancies. Since then, our model is consistently better than the GNN-based model, maintaining a margin of over $1$ \textsc{chrF++} score.
	\subsubsection{How Far Does Attention Look At} The Graph Transformer shows a strong capacity for processing complex and large graphs. We attribute the success to the global communication design, as it provides opportunities for direct communication in long distance. A natural and interesting question is how well the model makes use of this property. To answer this question, following \newcite{voita-etal-2019-analyzing}, we turn to study the attention distribution of each attention head. Specifically, we record the specific distance to which its maximum attention weight is assigned as attention distance. Fig. \ref{head} shows the averaged attention distance after we run our model on the development set of LDC2017T10. We can observe that nearly half of the attention heads have an average attention distance larger than $2$. The number of these far-sighted heads generally increases as layers go deeper. Interestingly, the longest-reaching head (layer1-head5) and the shortest-sighted head (layer1-head2) coexist in the very first layer, while the former has an average distance over 5.
	\section{Conclusions}
	In this paper, we presented the Graph Transformer, the first graph-to-sequence learning model based entirely on automatic attention. Different from previous recurrent models that require linearization of input graph and previous graph neural network models that restrict the direct message passing in the first-order neighborhood, our model enables global node-to-node communication. With the Graph Transformer, we achieve the new state-of-the-art on two typical graph-to-sequence generation tasks with four benchmark datasets.
	\bibliography{AAAI-CaiD.6741}

\begin{thebibliography}{}

\bibitem[\protect\citeauthoryear{Alberti \bgroup et al\mbox.\egroup
  }{2017}]{alberti2017syntaxnet}
Alberti, C.; Andor, D.; Bogatyy, I.; Collins, M.; Gillick, D.; Kong, L.; Koo,
  T.; Ma, J.; Omernick, M.; Petrov, S.; et~al.
\newblock 2017.
\newblock Syntaxnet models for the conll 2017 shared task.
\newblock {\em arXiv preprint arXiv:1703.04929}.

\bibitem[\protect\citeauthoryear{Bahdanau, Cho, and
  Bengio}{2015}]{bahdanau2015neural}
Bahdanau, D.; Cho, K.; and Bengio, Y.
\newblock 2015.
\newblock Neural machine translation by jointly learning to align and
  translate.
\newblock In {\em ICLR}.

\bibitem[\protect\citeauthoryear{Banarescu \bgroup et al\mbox.\egroup
  }{2013}]{banarescu2013abstract}
Banarescu, L.; Bonial, C.; Cai, S.; Georgescu, M.; Griffitt, K.; Hermjakob, U.;
  Knight, K.; Koehn, P.; Palmer, M.; and Schneider, N.
\newblock 2013.
\newblock Abstract meaning representation for sembanking.
\newblock In {\em Proceedings of the 7th Linguistic Annotation Workshop and
  Interoperability with Discourse},  178--186.

\bibitem[\protect\citeauthoryear{Bastings \bgroup et al\mbox.\egroup
  }{2017}]{bastings-etal-2017-graph}
Bastings, J.; Titov, I.; Aziz, W.; Marcheggiani, D.; and Sima{'}an, K.
\newblock 2017.
\newblock Graph convolutional encoders for syntax-aware neural machine
  translation.
\newblock In {\em EMNLP},  1957--1967.

\bibitem[\protect\citeauthoryear{Beck, Haffari, and
  Cohn}{2018}]{beck-etal-2018-graph}
Beck, D.; Haffari, G.; and Cohn, T.
\newblock 2018.
\newblock Graph-to-sequence learning using gated graph neural networks.
\newblock In {\em ACL},  273--283.

\bibitem[\protect\citeauthoryear{Cao and
  Clark}{2019}]{cao-clark-2019-factorising}
Cao, K., and Clark, S.
\newblock 2019.
\newblock Factorising {AMR} generation through syntax.
\newblock In {\em NAACL},  2157--2163.

\bibitem[\protect\citeauthoryear{Cho \bgroup et al\mbox.\egroup
  }{2014}]{cho2014learning}
Cho, K.; Van~Merri{\"e}nboer, B.; Gulcehre, C.; Bahdanau, D.; Bougares, F.;
  Schwenk, H.; and Bengio, Y.
\newblock 2014.
\newblock Learning phrase representations using rnn encoder-decoder for
  statistical machine translation.
\newblock {\em arXiv preprint arXiv:1406.1078}.

\bibitem[\protect\citeauthoryear{Dai \bgroup et al\mbox.\egroup
  }{2019}]{dai-etal-2019-transformer}
Dai, Z.; Yang, Z.; Yang, Y.; Carbonell, J.; Le, Q.; and Salakhutdinov, R.
\newblock 2019.
\newblock Transformer-{XL}: Attentive language models beyond a fixed-length
  context.
\newblock In {\em ACL},  2978--2988.

\bibitem[\protect\citeauthoryear{Damonte and
  Cohen}{2019}]{damonte-cohen-2019-structural}
Damonte, M., and Cohen, S.~B.
\newblock 2019.
\newblock Structural neural encoders for {AMR}-to-text generation.
\newblock In {\em NAACL},  3649--3658.

\bibitem[\protect\citeauthoryear{Denkowski and
  Lavie}{2014}]{denkowski:lavie:meteor-wmt:2014}
Denkowski, M., and Lavie, A.
\newblock 2014.
\newblock Meteor universal: Language specific translation evaluation for any
  target language.
\newblock In {\em Proceedings of the EACL 2014 Workshop on Statistical Machine
  Translation}.

\bibitem[\protect\citeauthoryear{Flanigan \bgroup et al\mbox.\egroup
  }{2016}]{flanigan-etal-2016-generation}
Flanigan, J.; Dyer, C.; Smith, N.~A.; and Carbonell, J.
\newblock 2016.
\newblock Generation from abstract meaning representation using tree
  transducers.
\newblock In {\em NAACL},  731--739.

\bibitem[\protect\citeauthoryear{Gu \bgroup et al\mbox.\egroup
  }{2016}]{gu-etal-2016-incorporating}
Gu, J.; Lu, Z.; Li, H.; and Li, V.~O.
\newblock 2016.
\newblock Incorporating copying mechanism in sequence-to-sequence learning.
\newblock In {\em ACL},  1631--1640.

\bibitem[\protect\citeauthoryear{Guo \bgroup et al\mbox.\egroup
  }{2019}]{guo2019densely}
Guo, Z.; Zhang, Y.; Teng, Z.; and Lu, W.
\newblock 2019.
\newblock Densely connected graph convolutional networks for graph-to-sequence
  learning.
\newblock {\em Transactions of the Association for Computational Linguistics}
  7:297--312.

\bibitem[\protect\citeauthoryear{Jones \bgroup et al\mbox.\egroup
  }{2012}]{jones-etal-2012-semantics}
Jones, B.; Andreas, J.; Bauer, D.; Hermann, K.~M.; and Knight, K.
\newblock 2012.
\newblock Semantics-based machine translation with hyperedge replacement
  grammars.
\newblock In {\em COLING},  1359--1376.

\bibitem[\protect\citeauthoryear{Kingma and Ba}{2014}]{kingma2014adam}
Kingma, D.~P., and Ba, J.
\newblock 2014.
\newblock Adam: A method for stochastic optimization.
\newblock {\em arXiv preprint arXiv:1412.6980}.

\bibitem[\protect\citeauthoryear{Kipf and Welling}{2017}]{kipf2017semi}
Kipf, T.~N., and Welling, M.
\newblock 2017.
\newblock Semi-supervised classification with graph convolutional networks.
\newblock In {\em ICLR}.

\bibitem[\protect\citeauthoryear{Koncel-Kedziorski \bgroup et al\mbox.\egroup
  }{2019}]{koncel-kedziorski-etal-2019-text}
Koncel-Kedziorski, R.; Bekal, D.; Luan, Y.; Lapata, M.; and Hajishirzi, H.
\newblock 2019.
\newblock {T}ext {G}eneration from {K}nowledge {G}raphs with {G}raph
  {T}ransformers.
\newblock In {\em NAACL},  2284--2293.

\bibitem[\protect\citeauthoryear{Konstas \bgroup et al\mbox.\egroup
  }{2017}]{konstas-etal-2017-neural}
Konstas, I.; Iyer, S.; Yatskar, M.; Choi, Y.; and Zettlemoyer, L.
\newblock 2017.
\newblock Neural {AMR}: Sequence-to-sequence models for parsing and generation.
\newblock In {\em ACL},  146--157.

\bibitem[\protect\citeauthoryear{Li \bgroup et al\mbox.\egroup
  }{2016}]{li2016gated}
Li, Y.; Tarlow, D.; Brockschmidt, M.; and Zemel, R.
\newblock 2016.
\newblock Gated graph sequence neural networks.
\newblock In {\em ICLR}.

\bibitem[\protect\citeauthoryear{Liu \bgroup et al\mbox.\egroup
  }{2015}]{liu-etal-2015-toward}
Liu, F.; Flanigan, J.; Thomson, S.; Sadeh, N.; and Smith, N.~A.
\newblock 2015.
\newblock Toward abstractive summarization using semantic representations.
\newblock In {\em NAACL},  1077--1086.

\bibitem[\protect\citeauthoryear{Papineni \bgroup et al\mbox.\egroup
  }{2002}]{papineni2002bleu}
Papineni, K.; Roukos, S.; Ward, T.; and Zhu, W.-J.
\newblock 2002.
\newblock Bleu: a method for automatic evaluation of machine translation.
\newblock In {\em ACL},  311--318.

\bibitem[\protect\citeauthoryear{Popovi{\'c}}{2017}]{popovic2017chrf++}
Popovi{\'c}, M.
\newblock 2017.
\newblock chrf++: words helping character n-grams.
\newblock In {\em Proceedings of the second conference on machine translation},
   612--618.

\bibitem[\protect\citeauthoryear{Pourdamghani, Knight, and
  Hermjakob}{2016}]{pourdamghani2016generating}
Pourdamghani, N.; Knight, K.; and Hermjakob, U.
\newblock 2016.
\newblock Generating english from abstract meaning representations.
\newblock In {\em INLG},  21--25.

\bibitem[\protect\citeauthoryear{See, Liu, and
  Manning}{2017}]{see-etal-2017-get}
See, A.; Liu, P.~J.; and Manning, C.~D.
\newblock 2017.
\newblock Get to the point: Summarization with pointer-generator networks.
\newblock In {\em ACL},  1073--1083.

\bibitem[\protect\citeauthoryear{Sennrich, Haddow, and
  Birch}{2016}]{sennrich-etal-2016-neural}
Sennrich, R.; Haddow, B.; and Birch, A.
\newblock 2016.
\newblock Neural machine translation of rare words with subword units.
\newblock In {\em ACL},  1715--1725.

\bibitem[\protect\citeauthoryear{Shaw, Uszkoreit, and
  Vaswani}{2018}]{shaw-etal-2018-self}
Shaw, P.; Uszkoreit, J.; and Vaswani, A.
\newblock 2018.
\newblock Self-attention with relative position representations.
\newblock In {\em NAACL},  464--468.

\bibitem[\protect\citeauthoryear{Song \bgroup et al\mbox.\egroup
  }{2016}]{song-etal-2016-amr}
Song, L.; Zhang, Y.; Peng, X.; Wang, Z.; and Gildea, D.
\newblock 2016.
\newblock {AMR}-to-text generation as a traveling salesman problem.
\newblock In {\em EMNLP},  2084--2089.

\bibitem[\protect\citeauthoryear{Song \bgroup et al\mbox.\egroup
  }{2017}]{song-etal-2017-amr}
Song, L.; Peng, X.; Zhang, Y.; Wang, Z.; and Gildea, D.
\newblock 2017.
\newblock {AMR}-to-text generation with synchronous node replacement grammar.
\newblock In {\em ACL},  7--13.

\bibitem[\protect\citeauthoryear{Song \bgroup et al\mbox.\egroup
  }{2018}]{song-etal-2018-graph}
Song, L.; Zhang, Y.; Wang, Z.; and Gildea, D.
\newblock 2018.
\newblock A graph-to-sequence model for {AMR}-to-text generation.
\newblock In {\em ACL},  1616--1626.

\bibitem[\protect\citeauthoryear{Srivastava \bgroup et al\mbox.\egroup
  }{2014}]{srivastava2014dropout}
Srivastava, N.; Hinton, G.; Krizhevsky, A.; Sutskever, I.; and Salakhutdinov,
  R.
\newblock 2014.
\newblock Dropout: a simple way to prevent neural networks from overfitting.
\newblock {\em The Journal of Machine Learning Research} 15(1):1929--1958.

\bibitem[\protect\citeauthoryear{Vaswani \bgroup et al\mbox.\egroup
  }{2017}]{vaswani2017attention}
Vaswani, A.; Shazeer, N.; Parmar, N.; Uszkoreit, J.; Jones, L.; Gomez, A.~N.;
  Kaiser, {\L}.; and Polosukhin, I.
\newblock 2017.
\newblock Attention is all you need.
\newblock In {\em NIPS},  5998--6008.

\bibitem[\protect\citeauthoryear{Voita \bgroup et al\mbox.\egroup
  }{2019}]{voita-etal-2019-analyzing}
Voita, E.; Talbot, D.; Moiseev, F.; Sennrich, R.; and Titov, I.
\newblock 2019.
\newblock Analyzing multi-head self-attention: Specialized heads do the heavy
  lifting, the rest can be pruned.
\newblock In {\em ACL},  5797--5808.

\bibitem[\protect\citeauthoryear{Xu \bgroup et al\mbox.\egroup
  }{2018}]{xu2018graph2seq}
Xu, K.; Wu, L.; Wang, Z.; Feng, Y.; Witbrock, M.; and Sheinin, V.
\newblock 2018.
\newblock Graph2seq: Graph to sequence learning with attention-based neural
  networks.
\newblock {\em arXiv preprint arXiv:1804.00823}.

\end{thebibliography}
	\bibliographystyle{aaai}
\end{document}